%% file: Arxiv.tex
\documentclass[10pt, twocolumn]{article}
\usepackage[a4paper, total={7in, 10in}]{geometry}
\usepackage{IEEEtrantools}
\usepackage{cite}
\usepackage[pdftex]{graphicx}
\usepackage{amsmath}
\usepackage{subcaption}
\usepackage{url}
\usepackage{xcolor}
\usepackage{enumitem}
\usepackage{setspace}
\definecolor{byzantine}{rgb}{0.74, 0.2, 0.64}

\date{}
\author{
 Szymon Buchaniec\\
  AGH University of Science and Technology \\
  Krakow, Poland \\ 
  \texttt{buchaniec@agh.edu.pl}
  \and
   Marek Gnatowski\\
    AGH University of Science and Technology \\
  Krakow, Poland \\ 
  \texttt{mgnatow@student.agh.edu.pl}
  \and
  Grzegorz Brus\\
   AGH University of Science and Technology \\
  Krakow, Poland \\ 
  \texttt{brus@agh.edu.pl}
  }
\title{\Large{\textbf{An Analysis of an Integrated Mathematical Modeling - Artificial Neural Network Approach for the Problems with a Limited Learning Dataset}}}
\begin{document}
\maketitle

\begin{spacing}{0.9}\footnotesize{\textbf{\input{0_abstract}}}\end{spacing}
\input{1_motivation}

\input{2_survey}
\input{3_idea}
\input{4_benchmark}

\input{6_results}
\input{7_conclusion}
\input{8_acknowledgements}
\bibliographystyle{IEEEtran}
\bibliography{bibliography}

\end{document}

%% file: 0_abstract.tex
\begin{abstract}
One of the most common and universal problems in science is to investigate a function. The prediction can be made by an Artificial Neural Network (ANN) or a mathematical model. Both approaches have their advantages and disadvantages. Mathematical models were sought as more trustworthy as their prediction is based on the laws of physics expressed in the form of mathematical equations. However, the majority of existing mathematical models include different empirical parameters, and both approaches inherit inevitable experimental errors. At the same time, the approximation of neural networks can reproduce the solution extremely well if fed with a sufficient amount of data. The difference is that an ANN requires big data to build its accurate approximation whereas a typical mathematical model needs just several data points to estimate an empirical constant. Therefore, the common problem that developer meet is the inaccuracy of mathematical models and artificial neural network. An another common challenge is the computational complexity of the mathematical models, or lack of data for a sufficient precision of the Artificial Neural Networks. In the presented paper those problems are addressed using the integration of a mathematical model with an artificial neural network. In the presented analysis, an ANN predicts just a part of the mathematical model and its weights and biases are adjusted based on the output of the mathematical model. The performance of Integrated Mathematical modeling - Artificial Neural Network (IMANN) is compared to a Dense Neural Network (DNN) with the use of the benchmarking functions. The obtained calculation results indicate that such an approach could lead to an increase of precision as well as limiting the data-set required for learning.
\end{abstract}

%% file: 1_motivation.tex
\section{Introduction}
\label{sec:motivation}
\begin{figure}[!t]
\begin{subfigure}{\columnwidth}
\centering
\includegraphics{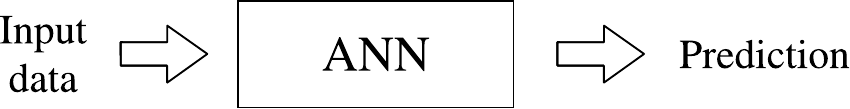}
\caption{}
\vspace{8pt}
\end{subfigure}
\begin{subfigure}{\columnwidth}
\centering
\includegraphics{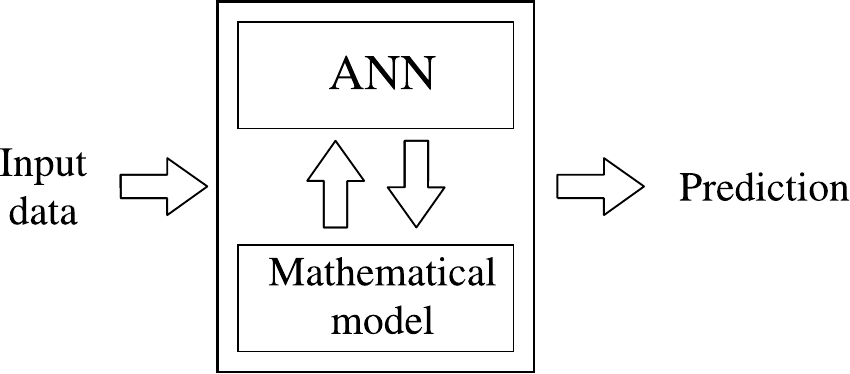}
\caption{}
\end{subfigure}
\caption{The comparison of two approaches to the prediction problem with use of an ANN (a) black box (b) combination of a mathematical model and ANN\label{fig:ideaBasic}}
\end{figure}
Predicting an output of a system is one of the most frequent tasks in fields of research in engineering, economics, or medicine. It may be a performance estimation of a device, cost analysis, or predictive medicine. Whenever it is possible, mathematical models are developed based on the laws of physics, observations and, unavoidably, assumptions. Mathematical models often can be inaccurate, incomplete, or very hard to formulate due to the gaps in knowledge. In such cases, to be able to predict a system's output, approximation methods are used. Computational models called Artificial Neural Networks have brought a vast improvement to predictions in many fields. An ANN can achieve excellent performance in function approximation, comparable to accurate mathematical models \cite{Nikzad2012}. For such high accuracy, an ANN needs a lot of data. In a typical learning process, more than twenty data points are needed per single input dimension. Obtaining enough data may be expensive and/or time-consuming, and in effect, unprofitable. Data numerousness is mostly dependent on the nonlinearity of a process one wants to predict. As an example of such a problem is a Solid Oxide Fuel Cell (SOFC) modeling, which is the main field of research of the authors of this publication. The SOFC's models are hard to be generalized over different types, have very complicated models because of transport and reaction kinetics phenomena. Obtaining data for one type of a SOFC, which is characterized by several parameters, may cost a few months of work. For such a problem, obtaining more than twenty data points per dimension would take years and therefore is infeasible  \cite{Brus:2015kt,Buchaniec:2019jo,Mozdzierz:tt}. The basic idea of the integration of an ANN with a mathematical model is presented in \figurename\ \ref{fig:ideaBasic}. 

The approach stems from the fact that the solution of every mathematical model is represented as a function or a set of functions. The method consists of determining the most uncertain parts of a mathematical model or lacking a theoretical description, and substitute them with a prediction of an ANN, instead of using the mathematical model or the ANN alone. With that being said, the work with the Interactive Mathematical modeling - Artificial Neural Network differs from the practice with a regular ANN. In a conventional procedure, a dataset is divided into training and validation data. An architecture, as well as a division of the data set, depends on the prediction's precision made on validation data. In addition to these steps, the IMANN requires to divide the problem into two parts. The mathematical model describes one of them, and the artificial neural network approximates the other. The decision regarding this division is a crucial part of working with IMANN, which affects the architecture of the network as well as the data. The main focus of the paper is to analyze the improvement of predictive accuracy under different level of integration of an ANN and a mathematical model.

This aim will be obtained by incorporating an artificial neural network to predict different parts of benchmark functions, which are regarded as a general representation of mathematical models. The ANN will be learned based on the mathematical model's output. A detailed description will be presented in section \ref{sec:bench}. In practice, it might be understood that the ANN prediction replaces only some equations in a model (or even just a part of an equation) and adapts to the system's behavior. As an example, let us consider a system of the equations in which one of the equations is replaced by the prediction done by an artificial neural network. The obtained approximation values of that function would be forced to fulfill all the equations in the system. If the prediction fails in doing so, there would be a discrepancy between measured and expected output during training. As a consequence, the ANN would be forced to improve its weights and biases until the system of equations is satisfied. Satisfying the system of equations ensures that the laws governing the system are included in the ANN. As we point later in this work such a replacement can benefit the accuracy and minimal dataset needed for an artificial neural network prediction. 



%% file: 2_survey.tex
\section{Literature review}
\label{sec:literature}
The problem with limited datasets is addressed frequently in the literature with many different approaches \cite{Andonie2010, Cataron2012, Shaikhina2017, Micieli2019}. One type of method is a data augmenting - the generation of a slightly different sample by modifying existing ones \cite{Simard2003}. Baird et al. used a combination of many graphical modifications to improve text recognition \cite{Baird1992}. Simard et al. proposed an improvement: the Tangent Prop method \cite{Simard1992}. In Tangent Prop, modified images were used to define the tangent vector, which was included in error estimation \cite{Simard1992}. Methods in which such vectors were used are still improved in recent works \cite{Rozsa2016, Lemley2017}. In the literature one can find a variety of methods, that can modify datasets, to improve ANN learning. A remarkably interesting approach, when dealing with two- and three-dimensional images is based on a persistent diagram (PD) technique. The PD changes the representation of the data to extract crucial characteristics \cite{Edelsbrunner2000} and uses as little information as possible to store them \cite{Adams2017}. Adcock et al. in \cite{Adcock2016} presented how persistent homology can improve machine learning. All the mentioned techniques are based on the idea to manipulate the dataset on which the ANN is trained.

An another method is to alter the ANN by including knowledge into its structure. A successful attempt to add the knowledge is made by a Knowledge-Based Artificial Neural Networks (KBANN) \cite{Towell1994}. The KBANN starts with some initial logic, which is transformed into the ANN \cite{Towell1994}. This ANN is then refined by using the standard backpropagation method \cite{Towell1994}. The KBANN utilizes a knowledge which is given by a symbolic representation in the form of logical formulas \cite{Towell1994}. In situations when knowledge is given by a functional representation i.e. containing variables, an interesting approach was presented by Su et al. in \cite{Su1992}. The authors proposed a new type of neural network - an Integrated Neural Network (INN) \cite{Su1992}. In the INN, an ANN is coupled with a mathematical model in such a way, that it learns how to bias the model's output, to improve concurrence with the modeled system \cite{Su1992}. The INN output consists of the sum of the model and ANN output \cite{Su1992}. A similar approach to improving the mathematical model was presented by Wang and Zhang in \cite{Wang1997}. They proposed an ANN, in which some of the neurons had their activation functions changed to the empirical functions \cite{Wang1997}. The ANN was used to alter the empirical model of the existing device in such a way, that it can be used for a different device. This idea led to a Neuro-Space Mapping (Neuro-SM) in which the functional model is a part of the ANN \cite{Bandler1999, Na2017}. Neuro-SM can be viewed as a model augmented by the ANN. In the Neuro-SM, the ANN maps the input and output of the model, but it does not interfere with the functional representation itself.

%% file: 3_idea.tex
\section{Methodology}
\label{sec:model}
\subsection{IMANN background}
\begin{figure}
\centering
\includegraphics{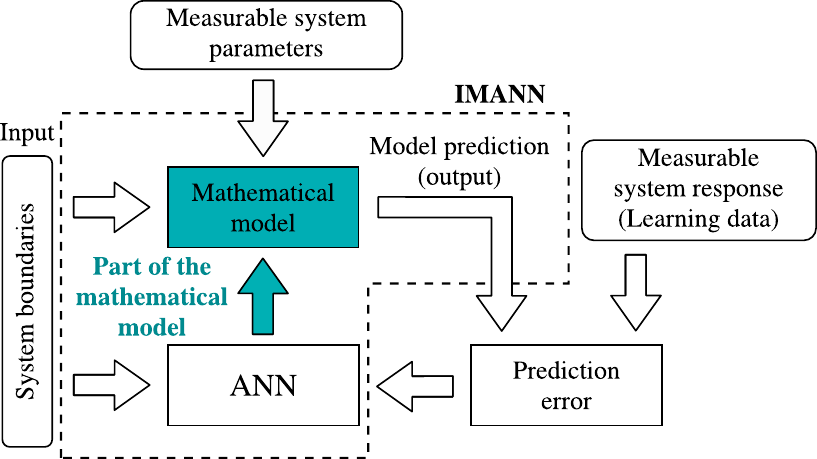}
\caption{The Integrated Mathematical Model - Artificial Neural Network block diagram \label{fig:ideaDetail}}
\end{figure}

\begin{figure}
\centering
\includegraphics{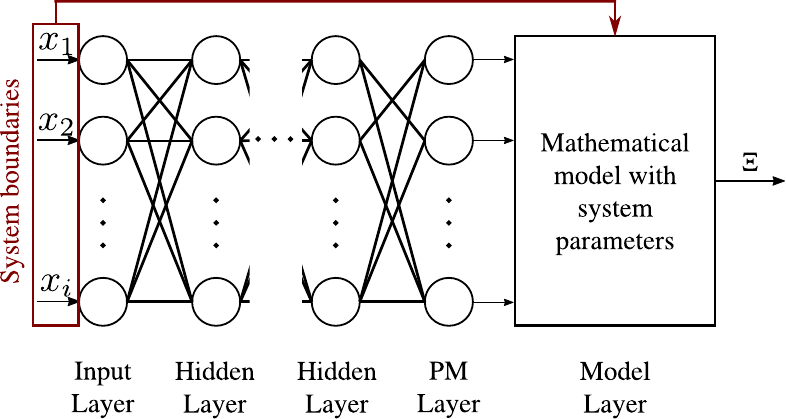}
\caption{Architecture schema of the IMANN \label{fig:imannImplementation}}
\end{figure}

The contribution of this paper is the in-depth analysis of an integrated Artificial Neural Network with a mathematical model. The integration is achieved by an interaction between the mathematical model and the ANN. The interaction is implemented by shifting a part of the mathematical model to be predicted by the ANN and teaching the ANN using a mathematical model's output errors. For instance, one can imagine that one of the equations in the model is replaced by ANN's prediction. In this approach, the model can be more flexible in predicting the shape of a function, and the ANN becomes more aware of the physics. If the equation predicted by the ANN fails to fulfill other equations in the system it will be reflected in the cost function, and thereby the ANN will be forced to improve its weights and biases.

\subsection{IMANN implementation}
System boundaries are supplied to the ANN, which predicts the assigned part of the mathematical model. System boundaries alongside with an ANN's output are provided to the mathematical model. The mathematical model computes the predicted system output. At the learning phase, the predicted output error is used to calculate the proper weights and biases of the ANN. The proposed approach to a prediction problem is graphically presented in \figurename\ \ref{fig:ideaDetail}. The selection of weights and biases is performed by an evolutionary algorithm. Every individual is a representation of one network, i.e. its weights and biases are in the form of a vector. The fitness function is an error measure of the mathematical model's output.

\label{sec:model}
\subsection{Model architecture}
The implementation of the IMANN has a feed-forward architecture consisting of fully connected layers with the last two layers called the part of the model layer (PM layer) and the model layer. In the input layer, the number of the neurons corresponds to the conditions in which the system is located. Further layers are standard hidden layers. The number of layers and the number of neurons contained therein are selected according to the complexity of the problem modeled by the network. The next layer is the mentioned PM layer, where the number of neurons corresponds to the number of replaced parts of the model. The output from that layer multiplied by the weights is directly entered into the model layer. The mathematical model, alongside with the system parameters, also receives all arguments, that were supplied to the network's input. The calculated mathematical model result is the final output of the IMANN. Schematically, the IMANN's architecture is presented in \figurename\ \ref{fig:imannImplementation}.

\subsection{Model learning process}
\label{sec:modelLearn}
To train the IMANN, the Covariance Matrix Adaptation - Evolution Strategy (CMA-ES) algorithm is chosen \cite{Hansen1996}. It is an effective and flexible algorithm, which is ideally suited to IMANN learning, where a problem taken under consideration can have different degrees of dimensionality. Here, the dimensionality of the problem is the number of weights and biases in the IMANN. To train the IMANN, the CMA-ES uses a vector made from the all network's weights and biases. The CMA-ES optimizes the vector values based on the error obtained on the training data. Details about the objective function will be discussed in the following section. Open source implementation of CMA-ES in Python was adopted for evolutionary weights adjustments \cite{hansen2019pycma}.

%% file: 4_benchmark.tex
\section{Benchmarking functions}
\label{sec:bench}
\begin{figure*}{}
\centering
\includegraphics{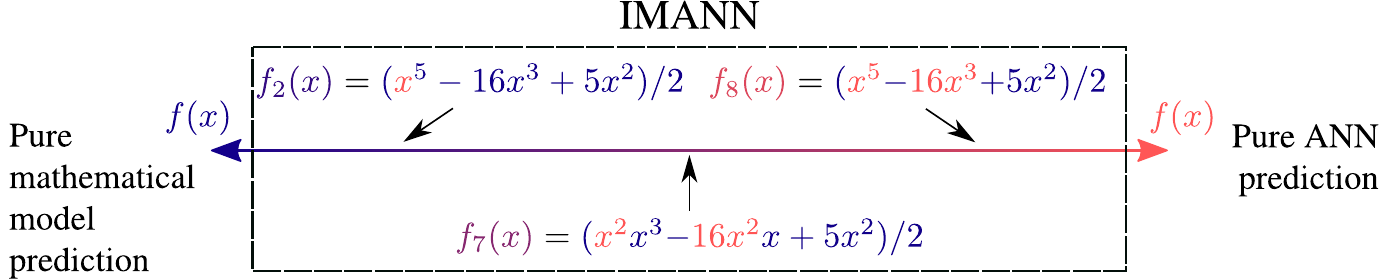}
\caption{Mathematical model and ANN prediction parts in case of polynomial function. IMANN extreme variations are pure mathematical model and pure ANN predictions \label{fig::shift}}{}
\end{figure*}
Every system's output, physical or theoretical, depends on its boundaries (inputs) and characteristics (adaptable parameters). Mathematically speaking, every system is a function. Mathematical models are functions that try to reflect real system behavior. Mathematical models are often based on assumptions and empirical parameters due to the gap in existing knowledge regarding the phenomena. The simplifications in problem formulations result in the discrepancy between the model and the real system outputs. The difference can dissolve only for hypothetical cases where system output is described by mathematical equations and well-defined. 

Every function can be treated as a system, and any arbitrary function can be treated as its mathematical model. 
The concept of IMANN strives to be reliable and applicable to any system: physical, economic, biological, or social, only when a part of this system can be represented in mathematical form. To be able to represent a wide range of possible applications, benchmark functions were employed as a representation of a system. 
The systems' inputs ($\boldsymbol{x}$) and measurable outputs ($\Xi(\boldsymbol{x})$) corresponds to the inputs and outputs of benchmarking functions. The ANN predicts a part of the system, and the mathematical model calculates the rest of it, treating the ANN's prediction as a parameter or one of the model's equations. The values calculated by the benchmarking functions will then represent the measurable system outputs. If the subfunction values predicted by ANN had the same value as calculated from the extracted part of the benchmark function, this would represent a perfect match between the IMANN and system output. If the real system output differs from the calculated one, the IMANN will be forced to improve weights and biases and predict the part of the system, that it is responsible for, again.

\subsection{Functions}
To test the IMANN, two benchmarking functions are used. One arbitrary polynomial function for a one-dimensional input and modified Rosenbrock function for two dimensions. The chosen polynomial function is given by formula:
\begin{equation}
f_P(x) = \frac{x^5 - 16x^3 + 5x^2}{2},
\end{equation}%
the $N$-dimensional modified Rosenbrock function is given by:
\begin{equation}\label{eq:modrosen}
f_R(\mathbf{\boldsymbol{x}}) = \sum_{i=1}^{N-1}\left[\left(x_{i+1}-x_i^2\right)^4 + \left(1-x_i\right)^4 \right].
\end{equation}
In the case of polynomial function, eight formulations of mathematical models are used, four with one subfunction and four with two subfunctions. The difference between the formulations lies only in the nonlinearity and number of the subfunctions. Model formulations with one subfunction are given by the following equations:
\begin{IEEEeqnarray}{rCl}\label{eq:polynomial}
\IEEEyesnumber
\IEEEyessubnumber*
f_1(x) &=& \frac{a(x)x^5 - 16x^3 + 5 x^2}{2},\label{eq:polya}\\
f_2(x) &=& \frac{\hat{a}(x)x^4 - 16x^3 + 5 x^2}{2}\label{eq:polyb},\\
f_3(x) &=& \frac{\tilde{a}(x)x^3 - 16x^3 + 5 x^2}{2}\label{eq:polyc},\\
f_4(x) &=& \frac{\bar{a}(x) - 16x^3 + 5 x^2}{2}\label{eq:polyd},\\
\end{IEEEeqnarray}%
where $a,\ \hat{a},\ \tilde{a}$ and $\bar{a}$ are subfunctions and $f_i$ is the $i$-th model function. For perfect match with the modeled benchmarking function, subfunctions should be functions of $x$ in the form:
\begin{IEEEeqnarray}{rCl}\label{eq:polyparams}
\IEEEyesnumber\IEEEyessubnumber*
a(x) &=& 1 \label{eq:polyparama} \\
\hat{a}(x) &=& x\\
\tilde{a}(x) &=& x^2\\
\bar{a}(x) &=& x^5.\label{eq:polyparamabar} 
\end{IEEEeqnarray}

The polynomial function's model formulations with two subfunctions are defined as:
\begin{IEEEeqnarray}{rCl}\label{eq:polynomial2}
\IEEEyesnumber
\IEEEyessubnumber*
f_5(x) &=& \frac{a(x)x^5 + b(x)x^3 + 5 x^2}{2}\label{eq:polye},\\
f_6(x) &=& \frac{\hat{a}(x)x^4 + \hat{b}(x)x^2 + 5 x^2}{2}\label{eq:polyf},\\
f_7(x) &=& \frac{\tilde{a}(x)x^3 + \tilde{b}(x)x + 5 x^2}{2}\label{eq:polyg},\\
f_8(x) &=& \frac{\bar{a}(x) + \bar{b}(x) + 5 x^2}{2}\label{eq:polyh},
\end{IEEEeqnarray}%
where $a,\ \hat{a},\ \tilde{a}$,\ $\bar{a},\ b,\ \hat{b},\ \tilde{b}$ and $\bar{b}$ are subfunctions. All model functions are defined in the domain $\Omega = \left\{x:x\in[-4,4]\right\}$. Ideally, subfunctions should be functions of $x$ in the form:
\begin{IEEEeqnarray}{rCl+rCl}\label{eq:polyparams2}
\IEEEyesnumber\IEEEyessubnumber*
a(x) &=& 1,\ & b(x) &=& -16, \label{eq:polyparama2} \\
\hat{a}(x) &=& x,\ & \hat{b}(x) &=& -16 x,\\
\tilde{a}(x) &=& x^2,\ & \tilde{b}(x) &=& -16 x^2,\\
\bar{a}(x) &=& x^5,\ & \bar{b}(x) &=& -16 x^3.\label{eq:polyparamabar2} 
\end{IEEEeqnarray}
The idea behind the problem formulation is presented in \figurename\ \ref{fig::shift}.

In the case of the two-dimensional modified Rosenbrock function, the model formulation is defined in $\Omega = \left\{(x,y): x \in[-1.4,1.6],\ y \in [-0.25, 3.75] \right\}$ and is given by:
\begin{equation}\label{eq:modrosenmodel}
f_{9}(x, y) = c_1^4(x,y) + c_2^4(x,y),
\end{equation}%
where $c_1$ and $c_2$ ideally are subfunctions of $x$ and $y$ in the form of:
\begin{IEEEeqnarray}{rCl+rCl}\label{eq:rosenparams}
c_1(x, y) &=& y-x^2,\ & c_2(x, y) &=&1-x.
\end{IEEEeqnarray}


The IMANN's ANN is responsible for predicting the value of all the subfunctions mentioned above and provides them into the model. The ANN is learned with the difference between the model output $\Xi(\boldsymbol{x}, \boldsymbol{w})$ and the data generated from a benchmarking function $\hat{\Xi}(\boldsymbol{x})$ in $n$ sample points, where $\boldsymbol{w}$ is the vector of weights and biases of a neural network. The learning process is performed by an evolutionary algorithm, here with the use of a CMA-ES library for Python as explained in section \ref{sec:modelLearn}. The vector $\boldsymbol{w}$, which fully describes one network, will be optimized based on the fitness value:
\begin{equation}
F(x) = \sum_{i=1}^{n}\left(\Xi\left(\boldsymbol{x}, \boldsymbol{w}\right)-\hat{\Xi}\left(\boldsymbol{x}\right) \right)^2.
\label{eq:squredErrorSum}
\end{equation}

%% file: 6_results.tex
\section{Results}
\label{sec:results}


\begin{figure}[t!]
\begin{subfigure}{\columnwidth}
\centering
\includegraphics{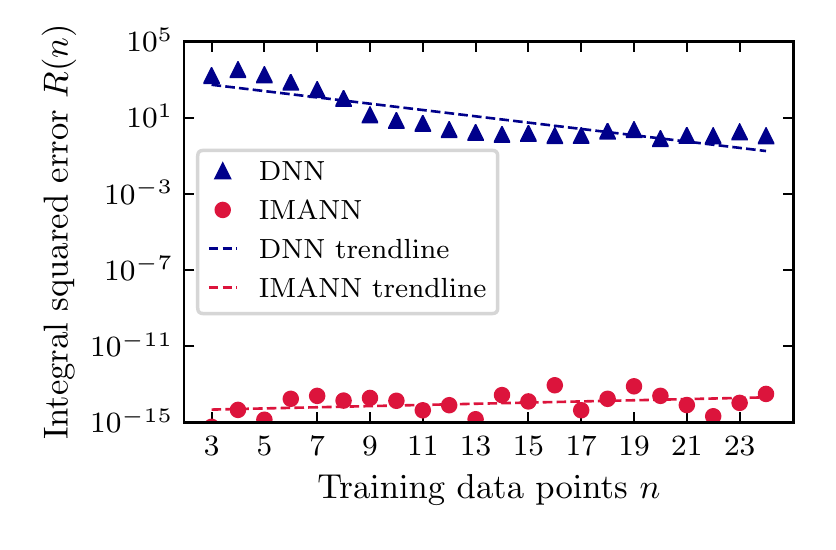}
\vspace{-1.2\baselineskip}
\caption{\label{fig:sub:const}}
\end{subfigure}
\begin{subfigure}{\columnwidth}
\centering
\includegraphics{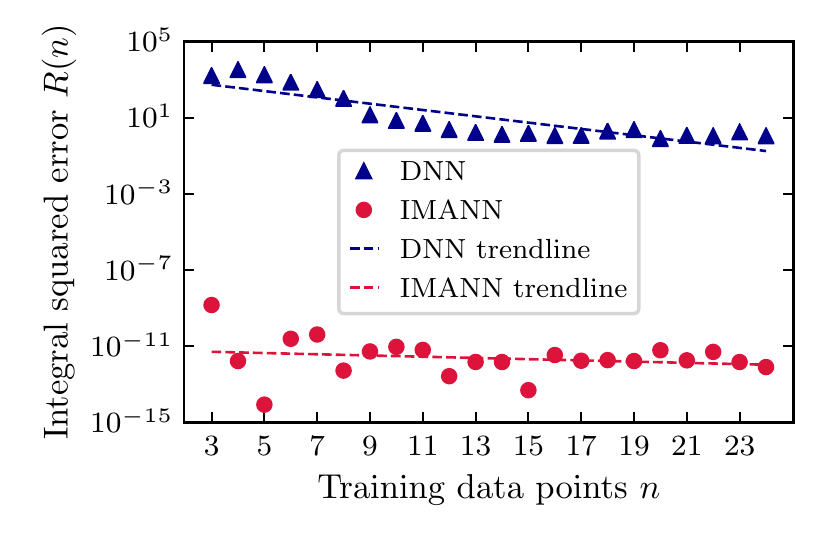}
\vspace{-1.2\baselineskip}
\caption{\label{fig:sub:linear}}
\end{subfigure}
\begin{subfigure}{\columnwidth}
\centering
\includegraphics{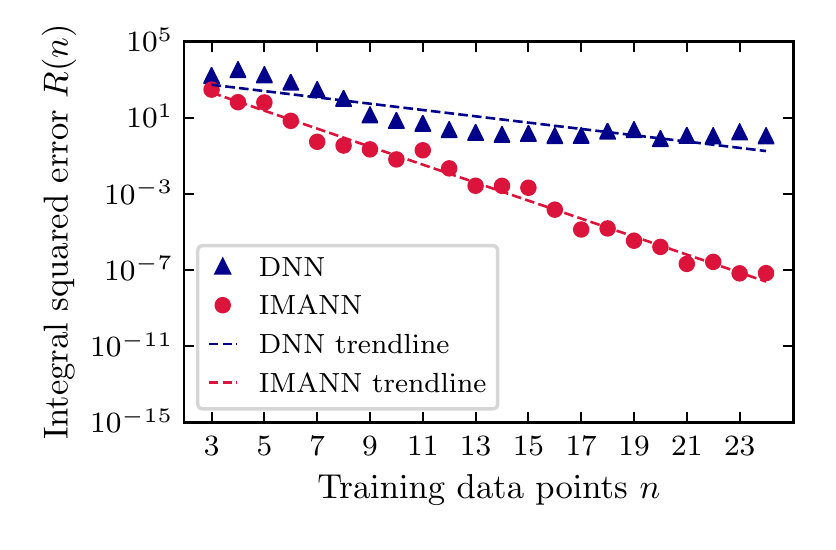}
\vspace{-1.2\baselineskip}
\caption{\label{fig:sub:nonlinear}}
\end{subfigure}
\begin{subfigure}{\columnwidth}
\centering
\includegraphics{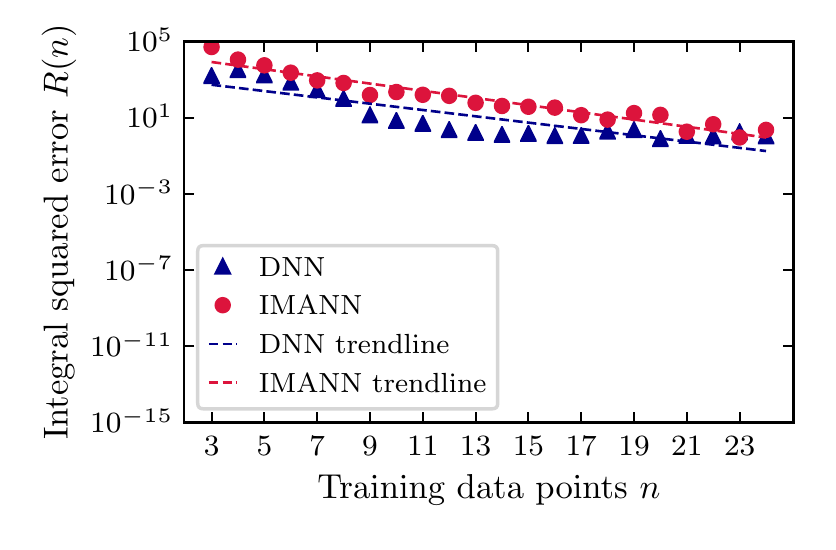}
\vspace{-1.2\baselineskip}
\caption{\label{fig:sub:full}}
\end{subfigure}
\caption{Squared error integral reduction with increase of dataset size for a problem of predicting a function (a) $f_1$ (b) $f_2$ (c) $f_3$ (d) $f_4$\label{fig:errorIntegral}}
\end{figure}

\begin{figure}[t!]
\begin{subfigure}{\columnwidth}
\centering
\includegraphics{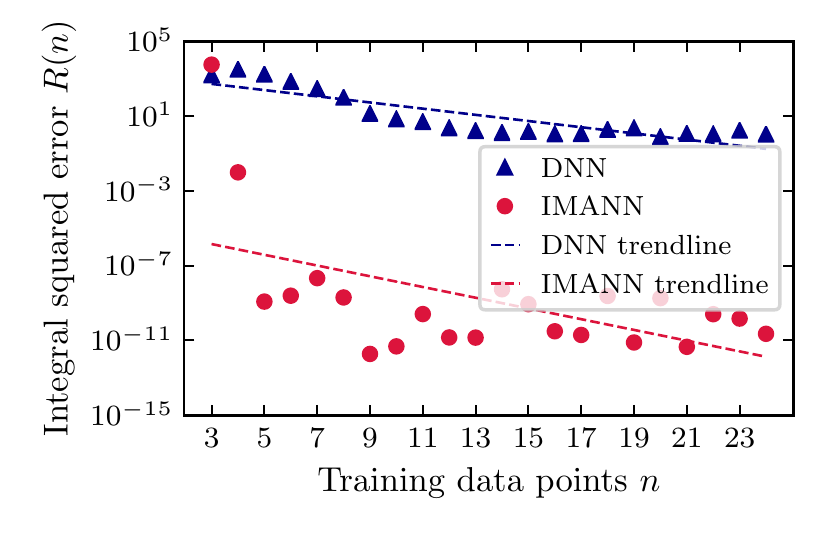}
\vspace{-1.2\baselineskip}
\caption{\label{fig:sub:const2}}
\end{subfigure}
\begin{subfigure}{\columnwidth}
\centering
\includegraphics{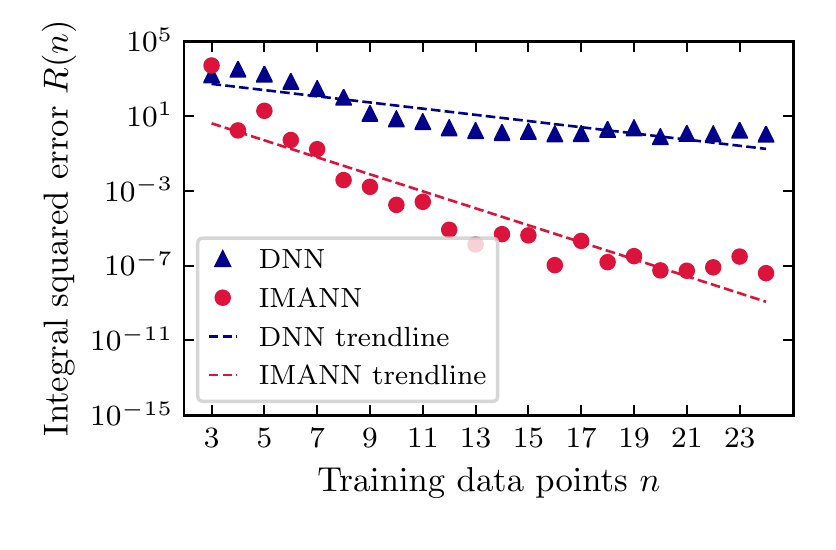}
\vspace{-1.2\baselineskip}
\caption{\label{fig:sub:linear2}}
\end{subfigure}
\begin{subfigure}{\columnwidth}
\centering
\includegraphics{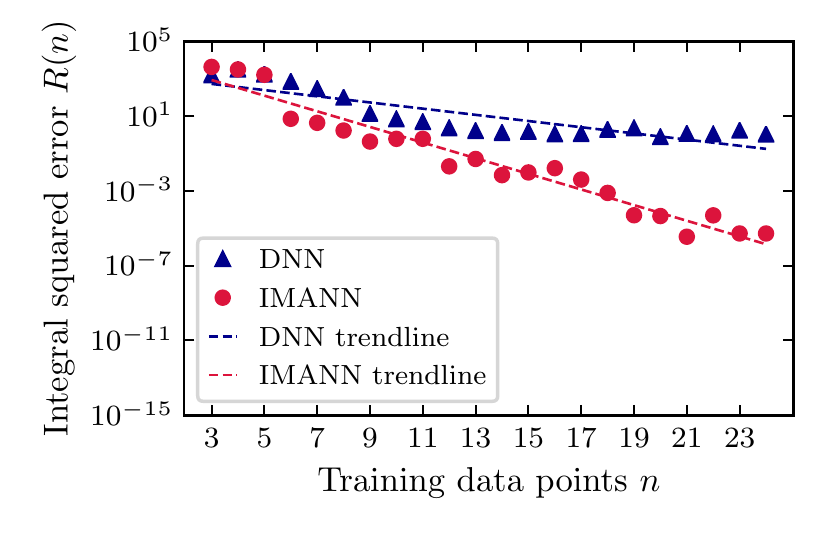}
\vspace{-1.2\baselineskip}
\caption{\label{fig:sub:nonlinear2}}
\end{subfigure}
\begin{subfigure}{\columnwidth}
\centering
\includegraphics{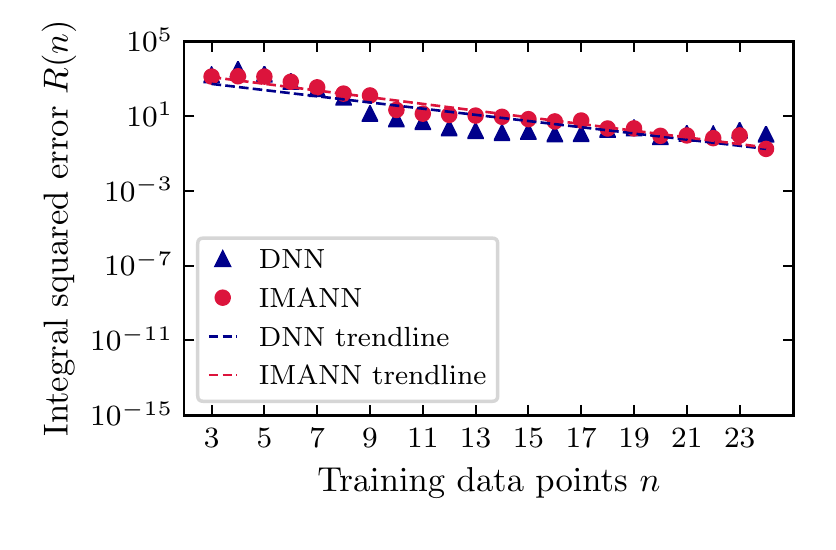}
\vspace{-1.2\baselineskip}
\caption{\label{fig:sub:full2}}
\end{subfigure}
\caption{Squared error integral reduction with increase of dataset size for a problem of predicting a function (a) $f_5$ (b) $f_6$ (c) $f_7$ (d) $f_8$\label{fig:errorIntegral2}}
\end{figure}

\begin{figure}[t!]
\includegraphics{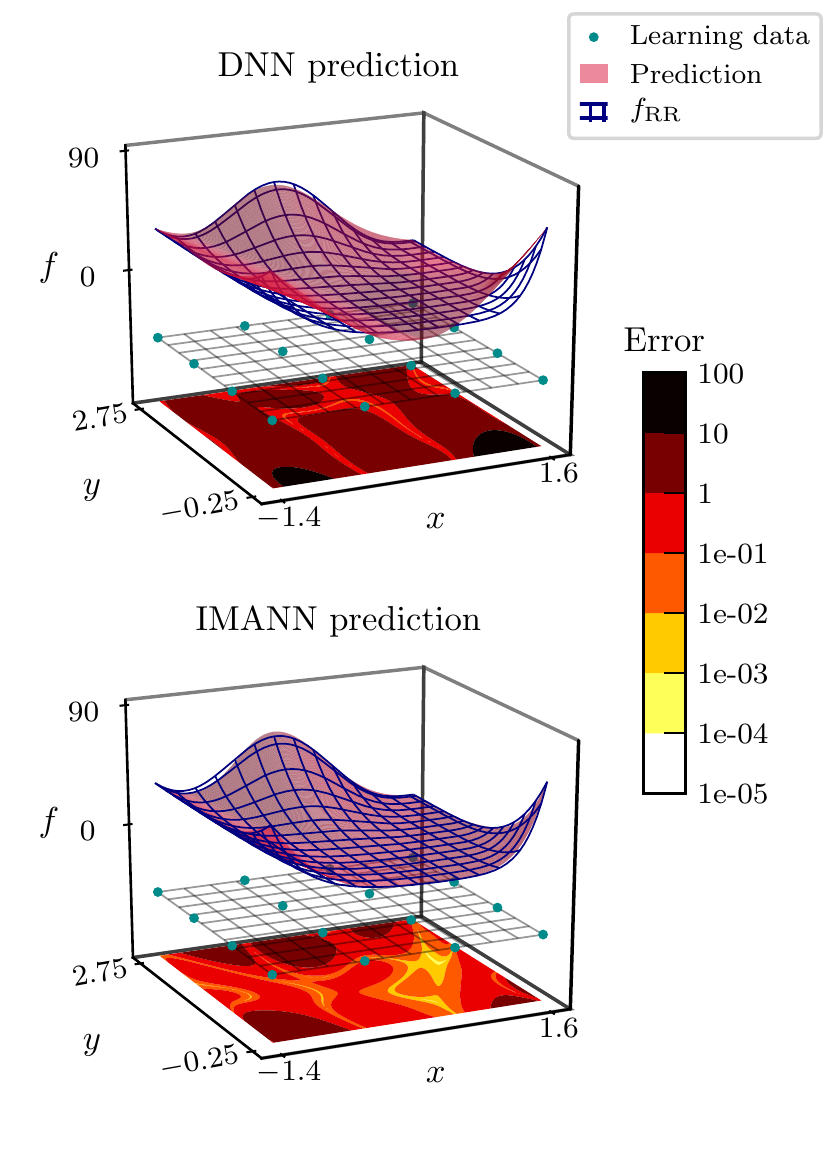}
\caption{Modified Rosenbrock function prediction with DNN and IMANN. Red surface corresponds to predicted value, wireframe to original function, blue points indicate learning dataset and colormap on the bottom indicates absolute error \label{fig::rosenerr}}
\end{figure}

\begin{figure}[t!]
\includegraphics{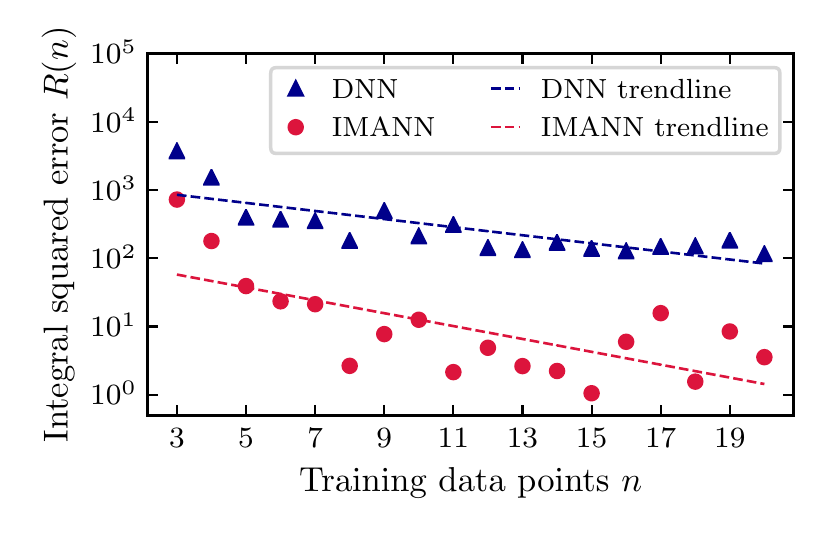}
\caption{Squared error integral reduction with increase of dataset size for $f_{9}$ \label{fig::roseninterr}}
\end{figure}

\begin{figure}[t!]
\includegraphics{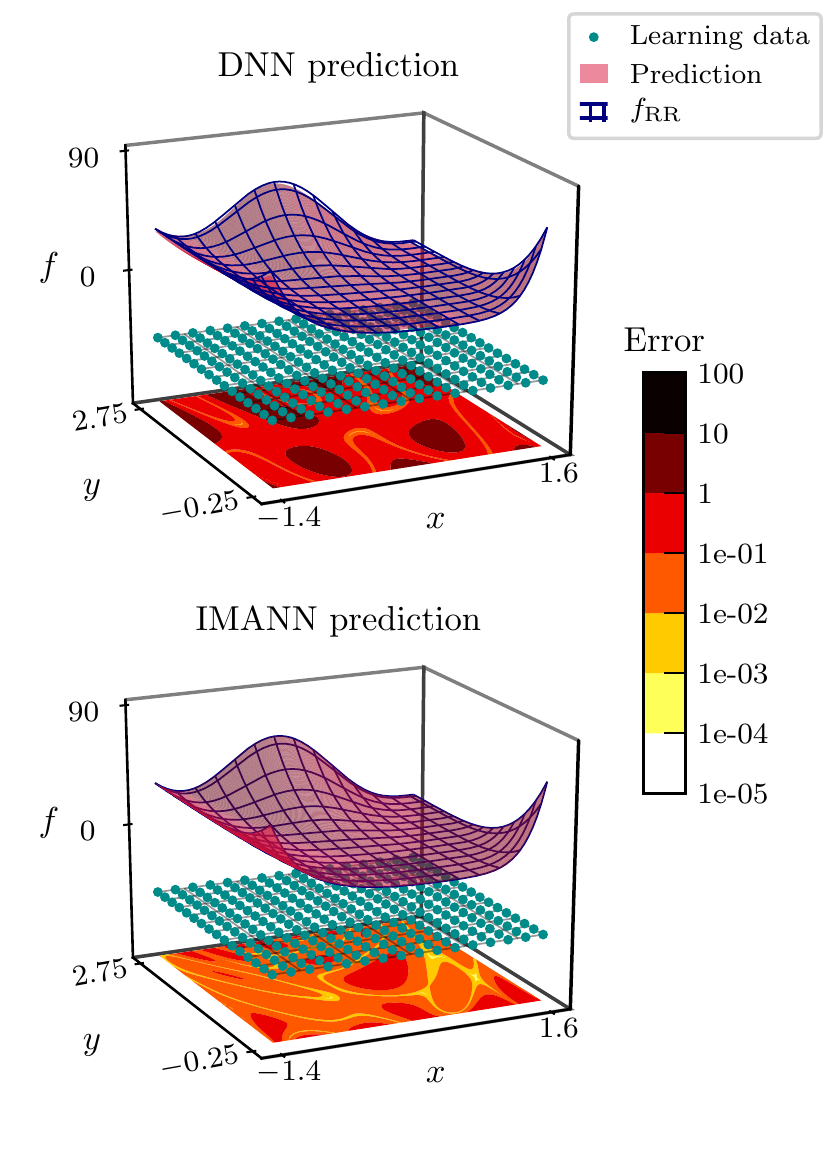}
\caption{Modified Rosenbrock function prediction with DNN and IMANN. Red surface corresponds to predicted value, wireframe to original function, blue points indicate learning dataset and colormap on the bottom indicates absolute error. DNN network architecture was similar to the IMANN's, i.e. 1-5-5-1 \label{fig::rosenerrdnn55}}
\end{figure}

\begin{figure}
\includegraphics{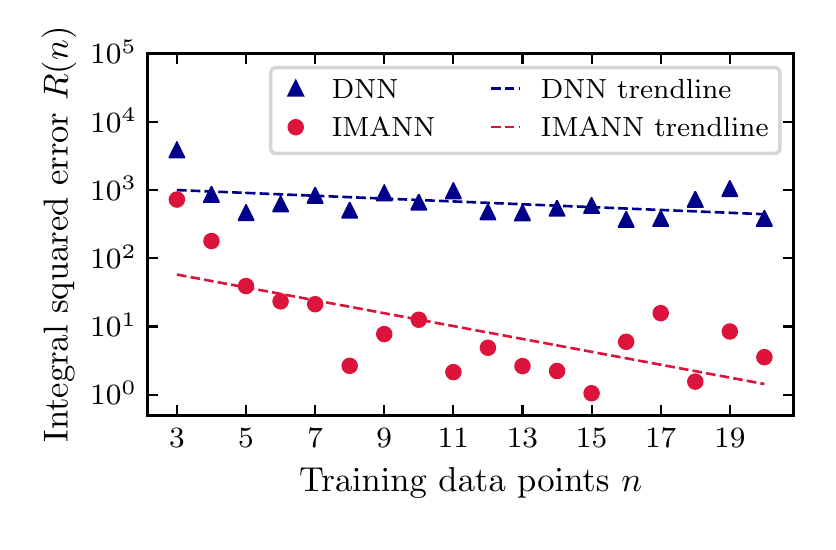}
\caption{Squared error integral reduction with increase of dataset size for $f_{9}$. DNN network architecture was similar to the IMANN's, i.e. 1-5-5-1 \label{fig::roseninterrdnn55}}
\end{figure}

To quantify the difference between the system's output and the predicted value, the integral of the squared error is used as an accuracy indicator:
\begin{equation}\label{eq:errorintegral}
R(\boldsymbol{x}, \boldsymbol{w}) = \int_\Omega{\sqrt{\left(\Xi\left(\boldsymbol{x}, \boldsymbol{w}\right)-\hat{\Xi}\left(\boldsymbol{x}\right)\right)^2}\mathrm{d}\boldsymbol{x}}.
\end{equation}
The integral in Eq. (\eqref{eq:errorintegral}) is computed with eighty point per dimension Gauss-{}Legendre quadrature. 
The IMANN is compared to DNN implemented with the use of the TensorFlow library \cite{tensorflow2015}. The DNN is learned in a standard way, treating the system as a black box. To neglect the stochastic error in the computations, 20 attempts were performed for the IMANN and DNN, and the best result, based on $R$ value, was taken. The term ANN will be used when referring to ANN part of IMANN and DNN for a typical ANN implementation.

The squared error integral value for the polynomial functions versus the number of the learning data points are presented in \figurename\ \ref{fig:errorIntegral}. The IMANN's architecture was 1-5-5-1 and DNN's 1-32-16-16-1, respectively. Subfigures from Fig. \ref{fig:sub:const}  to Fig. \ref{fig:sub:full} correspond to the task of predicting the functions from Eq. (\ref{eq:polya}) to Eq. (\ref{eq:polyd}) for both the IMANN and DNN. The IMANN divides tasks into two stages, firstly, the artificial neural network estimates Eq. (\ref{eq:polyparama}) to (\ref{eq:polyparamabar}), and secondly, the estimates are inserted into the mathematical models represented by functions from Eq. (\ref{eq:polya}) to Eq. (\ref{eq:polyd}). The performance comparisons between the IMANN and DNN are presented in \figurename\ \ref{fig:sub:const}-\ref{fig:sub:full}. As can be seen in \figurename\ \ref{fig:sub:const} the IMANN performs exceptionally well when only a constant is estimated by the ANN and the performance decay with the increase of non-linearity of the problem Figs. \ref{fig:sub:const}-\ref{fig:sub:nonlinear} with the uphold of the IMMAN advantage. With an increasing nonlinearity that is shifted from model to the ANN in the IMANN, the accuracy is getting closer to the DNN. When nonlinearity in the ANN is comparable to the overall model itself, the IMANN's prediction is worse than the DNN's (see \figurename\ \ref{fig:sub:full}). It is important to notice that the DNN has a much more complex architecture in comparison to the IMMAN. The increasing IMMAN architecture complexity is infeasible due to the utilization of the evolutionary algorithm for the adjustment of the weights. The dimensionality of the optimization of the fully connected network for the considered problem can be expressed with the following formula:
\begin{equation}
D = n_{\mathrm{in}}n_1+\sum_{i=2}^{m}n_{i-1}n_i + n_m n_{\mathrm{out}} + \sum_{i=1}^mn_i + 2n_{\mathrm{out}},
\end{equation}   
where $D$ is the dimensionality, $n_{\mathrm{in}}$ and $n_{\mathrm{out}}$ is the input and output dimensionality and $n_i$ is the number of neurons in the $i$-th hidden layer. For instance, the IMANN architecture for the polynomial prediction with one subfunction, the dimensionality of the optimization problem is equal to 47. 

Squared error integral value for polynomial functions versus number of learning data points while two subfunctions are predicted by ANN are presented in \figurename\ \ref{fig:errorIntegral2}. As before, IMANN's architecture was 1-5-5-1 and DNN's 1-32-16-16-1 respectively. Subfigures from Fig. \ref{fig:sub:const2} to Fig. \ref{fig:sub:full2} correspond to the task of predicting functions from Eq. \ref{eq:polye} to Eq. \ref{eq:polyh} for both IMANN and DNN. IMANN divides tasks into two stages, firstly, the artificial neural network estimates subfunctions Eq. \ref{eq:polyparama2} to Eq. \ref{eq:polyparamabar2}, and secondly, the estimates are inserted into the mathematical models represented by functions from Eq. \ref{eq:polye} to Eq. \ref{eq:polyh}. The performance comparisons between IMANN and DNN are presented in \figurename\ \ref{fig:sub:const2}-\ref{fig:sub:full2}. The increased difficulty and ambiguity of the problem, when the number of the subfunctions increases, causes the IMANN's prediction error to rise. The increase is especially significant when the linear part is being predicted by the ANN, and is equal to around seven orders of magnitude. Even with this increase, the IMANN performs five to ten orders of magnitude better than the DNN.

Figure \ref{fig::rosenerr} presents the approximation of the modified Rosenbrock function Eq. (\ref{eq:modrosen}) based on sixteen training data-points. The IMANN's architecture was 2-5-5-2 and the DNN's 2-32-32-16-1, respectively. The IMANN's ANN predicts subfunctions given in Eq. (\ref{eq:rosenparams}), and the estimates are inserted into the mathematical model represented by Eq. (\ref{eq:modrosenmodel}). The subfigures in \figurename\ \ref{fig::rosenerr} represent the prediction given by the DNN and IMANN. The contour maps at the bottom of each figure depict the error of the prediction as a function of system coordinates ($x$,$y$). The grid located above the contours indicates the training data, marked as dots. The prediction is displayed as a red surface, together with the original Rosenbrock function being a blue facade mesh. As can be seen in \figurename\ \ref{fig::rosenerr}, the IMANN achieved two order of magnitude higher prediction accuracy in the comparison to DNN when the networks were trained 256 data points. The precision of prediction as a function of training data-points for IMANN (2-5-5-2) and DNN (2-32-32-16-1) is presented in \figurename\ \ref{fig::roseninterr}. As can be seen in \figurename\ \ref{fig::roseninterr}, with the increase of the dataset the precision of both the DNN and IMMAN increases. However, starting from four data points the prediction precision of the IMANN is higher than the DNN's. Figure \ref{fig::rosenerrdnn55} presents the approximation of the modified Rosenbrock function based on 256 training data-points ($x$,$y$) for the same network architectures. The prediction of IMMAN is an order of magnitude better than the DNN's. The conclusion upholds when the same architectures of the IMANN (1-5-5-1) and DNN (1-5-5-1) are juxtaposed as it is presented in \figurename\ \ref{fig::roseninterrdnn55}. 

It can be concluded that by decreasing the load on the ANN part in the IMANN will result in the IMANN's prediction performance equal to the model. In our case, the model performance is perfect because we already know the form of the subfunctions. In real applications, finding even such a simple thing as the constant fitting parameters might be a problem. These generalized computations prove, that the IMANN can achieve higher performance than the DNN. The IMANN can improve mathematical model performance by modeling their over-simplified or missing parts. The obtained results indicate great potential in the integration of mathematical models and artificial neural networks.

%% file: 7_conclusion.tex
\section{Conclusions}
\label{sec:conclusion}
This paper presented an analysis regarding the integration of a mathematical model and an artificial neural network to limit the required dataset. The methodology can be applied to any system only if a part of it can be expressed in the form of mathematical equations. The combination of an artificial neural network and a mathematical model is interactive, which is expressed in the reinforcement of network weights adjustment based on the mathematical model misprediction. The Interactive Mathematical Model - Artificial Neural Network was employed to predict the values of several benchmark functions when given a different number of training data. The prediction of the IMANN was juxtaposed with the standard DNN network implemented in TensorFlow. The obtained results indicated that incorporating the mathematical model into an artificial neural network structure can be beneficial in terms of required data-sets, the precision of prediction, or the benefits in computational time. Replacing different parts of the model by artificial neural networks led us to the conclusion that the IMANN performs better when a more linear part of the model is replaced by the ANN prediction. This observation was not a surprise since it is in the core of the analyzed algorithm that it uses a synergy effect of the mathematical model and the artificial neural network. When an artificial neural network holds the primary function for the prediction, the synergy in the IMANN cannot be utilized any longer.

%% file: 8_acknowledgements.tex
\section*{Acknowledgements}
\label{sec:acknowledgements}
The presented research is the part of the Easy-to-Assemble Stack Type (EAST): Development of solid oxide fuel cell stack for the innovation in Polish energy sector project, carried out within the FIRST TEAM program (project number First TEAM/2016-1/3) of the Foundation for Polish Science, co-financed by the European Union under the European Regional Development Fund. The authors are grateful for the support. \\

This research made use of computational power provided by the PL-Grid Infrastructure.